\documentclass[sigconf]{acmart}
\AtBeginDocument{%
  \providecommand\BibTeX{{%
    \normalfont B\kern-0.5em{\scshape i\kern-0.25em b}\kern-0.8em\TeX}}}

\setcopyright{acmcopyright}
\copyrightyear{2022}
\acmYear{2022}
\acmDOI{XXXXXXX.XXXXXXX}

\acmConference[AIES '22]{2020 AAAI/ACM Conference on AI, Ethics, and Society}{August 1--3, 2022}{Oxford, England}
\acmBooktitle{AIES '22: 5th AAAI/ACM Conference on AI, Ethics, and Society} 
\acmPrice{15.00}
\acmISBN{978-1-4503-XXXX-X/18/06}

\begin{document}
\title{Towards a Responsible AI Development Lifecycle: Lessons From Information Security}

\author{Erick Galinkin}
\email{erick\_galinkin@rapid7.com}
\orcid{1234-5678-9012}
\affiliation{%
  \institution{Rapid7}
  \streetaddress{120 Causeway st}
  \city{Boston}
  \state{Massachusetts}
  \country{USA}
  \postcode{02114}
}

\renewcommand{\shortauthors}{Galinkin}

\begin{abstract}
Legislation and public sentiment throughout the world have promoted fairness metrics, explainability, and interpretability as prescriptions for the responsible development of ethical artificial intelligence systems.
Despite the importance of these three pillars in the foundation of the field, they can be challenging to operationalize and attempts to solve the problems in production environments often feel Sisyphean.
This difficulty stems from a number of factors: fairness metrics are computationally difficult to incorporate into training and rarely alleviate all of the harms perpetrated by these systems.
Interpretability and explainability can be gamed to appear fair, may inadvertently reduce the privacy of personal information contained in training data, and increase user confidence in predictions -- even when the explanations are wrong.
In this work, we propose a framework for responsibly developing artificial intelligence systems by incorporating lessons from the field of information security and the secure development lifecycle to overcome challenges associated with protecting users in adversarial settings.
In particular, we propose leveraging the concepts of threat modeling, design review, penetration testing, and incident response in the context of developing AI systems as ways to resolve shortcomings in the aforementioned methods.
\end{abstract}

\ccsdesc[500]{Security and privacy~Human and societal aspects of security and privacy}
\ccsdesc[500]{Social and professional topics~User characteristics}
\ccsdesc[300]{Social and professional topics~Computing / technology policy}

\keywords{Explainable Artificial Intelligence, Interpretable Machine Learning, Fairness, AI Ethics}

\maketitle

\section{Introduction}
Increasing adoption of artificial intelligence in public life has sparked tremendous interest in the fields of AI ethics, algorithmic fairness and bias, and model explainability and interpretability.
These ideas did not spring out of thin air, but rather are a response to difficult questions about when, where, and how it is appropriate to use artificial intelligence to perform tasks. 
Much of the extant literature has aimed to provide constructive movement in the direction of ensuring the principles of fairness, accountability, and transparency are upheld by machine learning algorithms.
For practitioners, there are three dominant approaches in AI ethics:
\begin{enumerate}
    \item Fairness metrics
    \item Interpretability
    \item Explainability
\end{enumerate}

Throughout this work, we use the term ``AI system'' to mean products and services that leverages artificial intelligence as a decision making component.
The term ``Responsible AI'' then is an AI system that is built with a notion of minimizing the potential harm.
The term ``harm'', used throughout the paper, we use in accordance with Crawford's~\cite{crawford2017trouble} use of the term to mean both allocative and representational harms.
Allocative harms are harms which result in an improper distribution of resources on the basis of group membership.
Representational harms are more difficult to quantify than allocative harms and result in the reinforcement of subordination of some group on the basis of identify, \textit{e.g.} race, gender, class, etc.

The topics of fairness, interpretability, and explainability are not merely of interest to the academic world.
The European Union has begun work on their ``AI Act''~\cite{eu2021proposal}, a law that seeks to legislate and harmonize regulations of technologies and products that leverage artificial intelligence.
In the United States, the National Institute of Standards and Technology has begun work on a risk management framework for artificial intelligence~\cite{nist2021rmf}, and a number of states have passed legislation regulating uses of artificial intelligence~\cite{ncsl2021legislation}.
Consequently, all at once, we are developing methods for determining and achieving fairness and explainability, implementing these methods in industry, and seeing regulation of the technologies that encourage or require those same methods.
Unfortunately, standardization of these topics is ongoing, there are no one-size-fits-all solutions, and there are significant methodological and computational hurdles to overcome.

We look to the field of information security as one potential model for success due to similarities between the two fields.
In particular, information security deals with a number of competing theories~\cite{moody2018toward} and standards~\cite{weidman2020nothing, sulistyowati2020comparative, roy2020high} that make it challenging to harmonize controls.
Moreover, information security, like ethical AI, aims to find heuristics, stopgaps, and proxies for computationally intractable problems~\cite{cohen1987computer, chess2004static} with important human impacts.
In information security, it is widely accepted that even in the best case for mitigation, vulnerabilities and compromises cannot be avoided entirely.
To this end, information security seeks to optimize mitigation, detection, and response.
In this work, we demonstrate how practitioners in ethical AI can use the framework of mitigation, detection, and response to operationalize fairness, interpretability, and explainabiltiy frameworks.

\section{Background and Related Works}
This work builds on research in economics, AI ethics, software engineering, and information security, drawing inspiration from Howard and Lipner's Security Development Lifecycle~\cite{howard2006security} and from the many ways that their work has been refined, implemented, and evolved in the software industry.
In this section, we provide background on fairness, interpretability, and explainability with an eye towards the insufficiency of existing methods.
Crucially, in the realm of interpretability and explainability, the presence of explanations increases user confidence in the predictions of the model, even when the explanations are incorrect~\cite{bansal2021does}.

\subsection{Fairness}
In the economics literature, there are a variety of fairness metrics that have been established. 
For many fairness metrics in the continuous case, the problems are rarely able to be solved efficiently~\cite{robertson1998cake} and for indivisible goods, envy-free allocation -- allocation where nobody would rather have someone else's good -- is \textsc{NP}-hard~\cite{lipton2004approximately}.
Kleinberg \textit{et al.}~\cite{kleinberg2017inherent} explore the COMPAS risk tool and show that for integer values, risk assignment is \textsc{NP}-Complete; although for non-integer values, the problem of fair risk assignment in polynomial-time remains open.
Fairness in classification was examined by Dwork \textit{et al.}~\cite{dwork2012fairness}, who developed fairness constraints in a classification context, identifying statistical parity as one way to determine fairness in classifiers.
Yona and Rothblum~\cite{yona2018probably} tackle the issue of generalization from training to test sets in machine learning by relaxing the notion of fairness to an approximation and demonstrating generalization of metric-fairness.
Like the aforementioned works, much of the literature focuses on the fairness of a single algorithm making classifications on groups of agents.

In the multi-agent setting, where motivations of individual agents may worsen the outcomes of other agents, the problem becomes even more difficult.
The work of Zhang and Shah~\cite{zhang2014fairness} attempts to resolve fairness in this multi-agent setting via linear programming and game theoretic approaches.
The game theoretic approach used tries to find a Nash equilibrium for the two-player zero-sum setting, a problem which is known to be \textsc{PPAD}-Complete~\cite{chen2006settling} and conjectured not to be in \textsc{P} unless \textsc{P} = \textsc{NP}.
This suggests that in general, any attempt at algorithmic fairness is a substantial computational problem on top of whatever problem we are aiming to solve.

In addition to computational difficulties, the work of Fazelpour and Lipton~\cite{fazelpour2020algorithmic} addresses shortcomings in the ideological foundations of formalizing fairness metrics by connecting the existing work to the political philosophy of ideal and non-ideal approaches.
As in much of the fair machine learning literature, ideal models in political theory imagine a world that is perfectly just. 
By using this as a target, we aim to measure -- and correct for -- the deviation from this ideal.
However, developing this fairness ideal in algorithmic settings necessitates comparison to other groups and consequently, a ``fair'' approach may actually be worse for all groups and yield new groups that need to be protected.
Further work by Dai \textit{et al.}~\cite{dai2021fair} shows that fairness allocations with narrow desiderata can lead to worse outcomes overall when issues like the intrinsic value of diversity~\cite{steel2021information} are not accounted for.
This suggests that because the term ``fairness'' is not well-defined, collaboration between developers of AI systems and social scientists or ethicists is important to ensure any metric for measuring fairness captures a problem-specific definition of the term.

\subsection{Interpretability}
Recent work on model interpretability has indicated that users find simpler models more trustworthy~\cite{schmidt2019quantifying}.
This is built on the definition of Lipton~\cite{lipton2018mythos} that presumes users are able to comprehend the entire model at once. 
However, the ability to interpret high-dimensional models is limited, even when those models are linear~\cite{molnar2020interpretable}.
This initially suggests that model interpretability limits the available models to low-dimensional linear models and short, individual decision trees.

Spurred by these notions, Generalized Linear Models and Generalized Additive Models have been developed and seek to be sufficiently robust to be useful in practice while retaining strong notions of human-interpretability.
These methods allow for linear and non-linear models that are inherently interpretable.
However, as Molnar notes~\cite{molnar2020interpretable}, high-dimensional models are inherently less interpretable even when those models are linear. 
Moreover, even the most interpretable models rely on assumptions about the stability of the data generation process and any violation of those assumptions renders interpretation of those weights invalid.

\subsection{Explainability}
Post-hoc explanations have proven very popular due to their intelligibility and their ability to be used with complex machine learning models, particularly neural networks.
Explainablity methods tend to be model agnostic and are more flexible than model-specific interpretation methods.
We refer readers interested in the technical details of these methods to other resources, such as the book by Molnar~\cite{molnar2020interpretable} or appropriate survey literature~\cite{carvalho2019machine}. 
In practice, explainability methods manifest in a variety of ways: 
\begin{enumerate}
    \item Partial Dependence Plots
    \item Individual Conditional Explanations
    \item Accumulated Local Effects
    \item Feature Interaction
    \item Feature Importance
    \item Global Surrogates
    \item Local Surrogates
    \item Shapley values
    \item Counterfactual Explanations
    \item Adversarial Examples
    \item Attention Layer Visualization
\end{enumerate}

The above methods can be broadly grouped into two buckets: global explanations and local explanations.
Global explanations seek to provide overall model interpretability for models that are otherwise difficult to understand.
These methods will demonstrate, for example, how certain features are weighted more heavily than others or show how correlation between variables can cause a particular prediction.
Local methods, on the other hand, purport to provide explanations for individual predictions. 
The most popular among these are LIME~\cite{ribeiro2016should}, GradCAM~\cite{selvaraju2017grad}, and SHAP~\cite{lundberg2017unified}, which leverage local surrogate models, gradient-based localization, and Shapley values respectively to foster explanations. 
In response to their popularity, the robustness of these methods have been investigated.
Slack \textit{et al.}~\cite{slack2020fooling} demonstrated that these methods do not work well in an adversarial setting -- that is, they can be fooled by a modeler who wishes to provide convincing explanations that appear innocuous while maintaining a biased classifier.
Further work by Agarwal \textit{et al.}~\cite{agarwal2021towards} attempts to establish foundations for robustness in explanation methods, finding that there are some robustness guarantees for some methods, but those guarantees are subject to variance in the perturbations and gradients.
Beyond these issues, substantial critiques have been leveraged against the use of Shapley values for feature importance~\cite{kumar2020problems} based on their inconsistency across distributions and the lack of a normative human evaluation for the values~\cite{kaur2020interpreting}.

Counterfactual explanations offer a particularly useful line of explanation, effectively answering the question: ``what would need to be different to get a different outcome?''
Humans desire counterfactual explanations, since they provide a direction to create a different outcome in the future~\cite{pearl2018book}.
As an example, when a person applies for a bank loan and is denied on the basis of their credit score, they expect a counterfactual explanation that says what factors, specifically, contributed to the denial and would need to improve in order to approve the loan.
Though metacognition -- thinking about thinking -- has been studied in computer science, and particularly in cognitive architectures~\cite{marshall1999metacat}, recent attempts have been made~\cite{kumar2021meta, babu2012meta, vilone2021notions} toward a metacognition for explaining difficult to interpret models, largely in the mold of providing counterfactual explanations.
However, to date, counterfactal explanations and artificial metacognition have not developed sufficiently to allow for their use.

\subsection{Attacks on AI systems}
There is a deep connection between security and fairness in machine learning systems.
Aside from clear connections like the link between differential privacy and fairness in classification~\cite{dwork2012fairness}, techniques like adversarial examples~\cite{szegedy2013intriguing} -- inputs to models that are similar to humans but are perturbed to cause misclassification -- can be used to evaluate the robustness of model fairness, interpretability, and explainability.
Adjacent to our taxonomy of allocative and representational harms, we also have a taxonomy of harms that our model can perpetrate against users and third parties: one, the harms caused by the system itself, including the aforementioned allocative and representational harms; two, the harms caused to users by other users of the system.
The first case is well-studied, though strategies for renumerating and redressing uncovered harms outside of calibration primarily prescribe putting human-in-the-loop or mandating explanations for a human gatekeeper.
The harms caused to users by other users of the system tend to align more closely with attacks on AI systems, which we provide a high-level overview of below and refer readers to surveys on attacks in machine learning~\cite{pitropakis2019taxonomy} and threats to privacy in machine learning~\cite{al2019privacy} for additional details.
These user-on-user harms largely align with four overarching categories: 
\begin{enumerate}
    \item Classification-level attacks
    \item Model-level attacks
    \item System-level attacks
    \item Privacy attacks
\end{enumerate}

Classification-level attacks are those attacks which seek to cause misclassification.
These attacks include adversarial examples in images, but also distinct techniques like ``Bad Characters''~\cite{boucher2022bad} that use imperceptible characters to bypass text content filters.
Essentially, these attacks allow one user to harm another by causing an input to be misclassified without altering the model, data, or anything else.
These attacks would also include attacks like the one used against Tesla's Traffic Aware Cruise Control~\cite{povolny2020model} where a malicious individual could easily modify a 35 mph speed limit sign with a small piece of black tape and cause the model to incorrectly classify the sign as an 85 mph speed limit sign.

Model-level attacks differ from classification-level attacks in that they alter the model itself.
The most common example of this is a poisoning attack -- an attack in which the training data of the model are altered to cause consistent misclassification.
This often requires access to the model or the data itself, making the attack challenging.
However, in the online setting, a number of online data poisoning attacks~\cite{zhang2020online,kloft2007poisoning} have been demonstrated to great effect.
A malicious user then, could poison the model and cause problems for all users.

System-level attacks are intend not to simply affect the predictions of the model, but rather damage the system itself.
An example here is that of sponge examples~\cite{shumailov2021sponge}, model inputs that are generated to maximize energy consumption and inference time to degrade the functionality of the system.
This can also include exploitation of conventional vulnerabilities which could allow for tampering with model inputs or outputs to harm users.

Privacy attacks include membership inference~\cite{shokri2017membership,choquette2021label} and model inversion~\cite{fredrikson2015model,carlini2021extracting}.
Membership inference attacks seek to identify whether or not individuals are present in the training data of a model, potentially damaging user privacy.
Model inversion then, is a step further. 
Rather than ask whether or not a user's data is present in the training data of the model, model inversion seeks to extract training data directly from the model -- a phenomenon that has been observed in generative models.
Both of these attacks can facilitate harms to users and are within the purview of responsible AI to limit.

\section{Adapting the Secure Software Development Lifecycle to Artificial Intelligence} \label{sec:adapting}
As discussed, attempts to satisfy fairness criteria can be limiting from a computational perspective.
Within information security, there is a notion of formal verification~\cite{ringer2020qed}, a computationally intensive process of ensuring that under any input, the program behaves as expected. 
This leads to more reliable software that is less prone to exploitable bugs.
Note however, that the mission statement of formal verification -- designing a program that halts when a bug is detected -- is undecidable because it is exactly the halting problem.
This has led to extensive work in both automated and interactive verification to overcome this theoretical barrier by solving subproblems, approximations of the problem, or writing domain-specific automation. 
In many cases, formal verification for software is a larger engineering effort than the software project itself and as a result, most software is not formally verified.
How then, do we ensure that software is not riddled with exploitable bugs?
In general, the presence of exploitable bugs in software~\cite{wysopal2006art,dowd2006art} is reduced through a number of steps in the secure software development lifecycle.
For our purposes, we identify analogies between ethical AI development and the following:
\begin{enumerate}
    \item Design Review
    \item Threat Modeling
    \item Penetration Testing
\end{enumerate}

These principles reduce risk that may be introduced in software development and produce more robust code without the overhead of formal verification methods.
In ethical artificial intelligence, we also seek to reduce the risk of negative outcomes and discrimination.
As such, we adapt these secure software development lifecycle principles to ethical AI.
One point of disagreement in the security community that may be reflected here is whether to perform threat modeling ahead of design review. 
The idea of performing threat modeling first is to provide a thorough view of the threats so that the risks uncovered in design review are threat-centric.
We follow the convention of performing design review ahead of threat modeling based on the rationale that defining the threats for a system that has not yet been designed makes the scope too broad to be useful.
We note that both approaches are valid and can be tailored to fit the maturity and preferences of the organization.

\subsection{Design Review} \label{sec:design_review}
In information security, a design review looks at the system under development and assesses the architecture, design, operations, and their associated risks~\cite{dowd2006art} allowing for implementation of systems-level security controls such as authentication, encryption, logging, and validation.
When developing AI systems, a similar sort of design review should be conducted, with a view toward AI risks.
This means that during the design review process, we should explore questions like:
\begin{itemize}
    \item How can we check for distribution drift?
    \item Are we logging model queries in a way that allows us to find reported bad behaviors?
    \item What features do we input to the model, and do they introduce potential issues?
    \item Are there other data sources we should be incorporating into this model?
    \item What actions, if any, are taken automatically as a result of model predictions?
\end{itemize}

This step provides a system-level view of how data goes into and predictions come out of the system and is ideally conducted before the system is deployed.
The idea, at the design review step, is to identify data flows and consider how the system could be refactored or rearchitected to avoid potential risks.
Things like data pre-processing or calibration~\cite{barocas2017fairness} should be discussed at this step, and if they are not needed or not sufficient, there should be documentation as to why they are omitted.
This goes beyond the actual model and training pipeline to include where data is derived from, what additional data is collected, where predictions and logs are stored, and other system-level issues.

An important part of the design review process is a discussion of how data related to the system is generated, processed, and stored.
This is an important part of the system that is often viewed through a lens of privacy and policy, but not always with a view of how to responsibly manage data.
While data management and mismanagement can cause one to run afoul of data privacy legislation, there are a variety of personal data misuses~\cite{kroger2021data} that can cause harm.
This means that the privacy of data \textit{per se} is not the entirety of the discussion, but how the data moves through the system to become a classification needs to be uncovered.
An investigation into this requires analysis of all data used in predictions, whether these are raw data, proxy features that stand in for data that is not directly available, or transformed features like those yielded from principle component analysis.

\subsection{Threat Modeling} \label{sec:threat_modeling}
Threat modeling is the phase of the development process that aims to predict the threats that a system may face~\cite{shostack2014threat}.
Akin to how one might imagine ways to secure a home by evaluating the locks, windows, and entrances to their home, threat modeling seeks to evaluate how attackers may gain entry to a system.
Since AI systems are software, the security threat modeling conducted should incorporate those systems.
By analogy, we want to think not only of threats to our system, but how our system could pose a risk to users.
This comes in two forms: malicious users of our system harming other users, and harms that our system could hypothetically cause.

When it comes to harming other users, we look to AI security and data privacy for potential harms~\cite{kroger2021data}.
Essentially, we must assess if users are fully independent and if not, the ways in which one user could potentially harm another.
As an example, malicious users could extract training data from trained models or infer individuals membership in the training data~\cite{carlini2021extracting,choquette2021label} which could then be used to harm those individuals privacy.
Another example is malicious users conducting data poisoning attacks~\cite{ahmed2021threats}, particularly for online machine learning systems~\cite{zhang2020online} that might lead to bad outcomes for other users.
This is one way that AI security directly influences AI ethics.

On the other hand, enumerating ways in which a system using AI could harm users is also critical.
Some harms may be expected: a self-driving car that does not recognize a pedestrian~\cite{shepardson2021us}, a discriminatory bail-setting algorithm~\cite{angwin2016machine}, an image cropping algorithm suffering from the ``male gaze''~\cite{birhane2022auditing}.
However, other harms could rear their head.
For example, the EMBER malware (malicious software) dataset~\cite{anderson2018ember} includes a large number of features for Windows Portable Executable files, including the language of the system the malware was compiled on.
One could conclude, based on the command and control infrastructure and the compilation language of the malware, that the presence of Chinese language is indicative of maliciousness and correspondingly restrict access to Chinese language websites.
One harm this could introduce, however, is inadvertent discrimination against Chinese-speaking users who may wish to visit legitimate webpages or run legitimate software.
Ultimately, we may conclude that the benefit of deploying the system outweighs the risk -- but identifying this possible harm is still an important part of the threat modeling process that we will revisit in our section on Incident Response.

\subsection{Penetration Testing} \label{sec:pen_test}
The concept of a penetration test is simple -- a trusted individual or team with adversarial skills seeks to find weaknesses in a system in accordance with the same techniques an attacker would use.
In the context of developing ethical AI systems, a ``penetration test'' then approaches our AI system with the same tools and intent as a malicious actor.
This test should evaluate an attacker's ability to harm the system, harm the users of a system, and also uncover harms latent in the system.
Much like the Twitter Algorithmic Bias Bug Bounty~\cite{chowdhury2021introducing}, we can and should directly evaluate our algorithms from an adversarial perspective, even if only internally.
Though the term penetration testing has a particular meaning in the information security context, we use it here to refer to the use of adversarial techniques to uncover potential harms in AI systems.
Additionally, we eschew the phrase ``algorithmic bias assessment'' since bias is only one potential cause for harm and we seek to use a more task-oriented term.

Conducting these sort of assessments require both AI security skills and sociotechnical knowledge.
As of 2021, only 3 out of 28 organizations surveyed conducted security assessments on their machine learning systems~\cite{kumar2020adversarial}, suggesting that many organizations are not currently well-equipped to evaluate these vulnerabilities and would need to cultivate teams capable of performing algorithmic harm assessments.
Utilities like Counterfit~\cite{pearce2021ai} and PrivacyRaven~\cite{hussain2020privacyraven} have lowered the barrier to entry for security professionals to use adversarial examples and membership inference attacks on machine learning models, but many organizations still do not assess their machine learning security.
These same utilities are critical to conducting these assessments against models.
Additionally, simple tactics like using so-called beauty filters can also demonstrate bias in machine learning systems~\cite{fingas2021twitter}.
In order to devise new tactics to target these algorithms, AI assessors need to understand both the technical and social factors included in these systems.
Importantly, the act of testing these systems assists us not only in identifying potential harms but also in assessing the robustness of our system.

Another key to penetration testing is the need to test the full system as deployed.
Since the algorithm is not deployed in a vacuum, there may be feature engineering, allow and block-listing, preprocessing, post-processing, and other steps that could allow problems to creep into the system.
Many so-called ``AI systems'' are not single algorithms deployed behind an API, but are instead a tapestry of data engineering, multiple algorithms, and post-processing systems. 
In some cases, an algorithm may be biased against a particular group, but some calibration~\cite{barocas2017fairness} in a post-processing system corrects for the identified issue.
In other cases, the added complexity of the overall system may actually amplify small changes to inputs and cause a larger effect that one might observe on the individual algorithm.

\section{Incident Response} \label{sec:incident}
An often overlooked discussion is how to deal with a harm perpetrated by an AI system once it is identified.
In the field of information security, there is the concept of a breach -- a successful intrusion by an attacker into our system -- and when this occurs, we begin the incident response process.
Typically an incident response process occurs alongside execution of a business continuity plan, a predefined plan for how to continue execution when there is a security event or natural disaster. 
The incident response process involves eliminating the attacker's access to systems, patching vulnerabilities that were exploited, and taking steps to ensure that the attacker does not get back in.
Similarly, we should be prepared in the field of AI to respond to events where our system creates or perpetuates harm.

There are a number of ways harms can be identified even after design review, threat modeling, and penetration testing such as through a bias bounty, a news report, or a user reporting that they have been harmed.
Once the existence of a harm is identified, the work of incident response begins with identifying what the actual harm is.
This can be an acute damage or harm to an individual, a systemic bias problem, or the potential for a third-party to harm other users of the system.
A self-driving car that strikes a pedestrian is a commonly-used example because the harm is clear: there exists a configuration of vehicles, pedestrians, and other distractions such that the vehicle does not stop before a pedestrian is struck.
Other harms, such as bias against racial and gender minorities as observed in the cases of COMPAS~\cite{angwin2016machine} and Amazon's hiring algorithm~\cite{dastin2018amazon} are less obvious until we conduct research into exactly what harms occurred.
Whether the harm identified is an acute damage to an individual or an ongoing systemic harm, we must take immediate action to:
\begin{enumerate}
    \item Continue operations if possible
    \item Perform root cause analysis
    \item Remediate the harms caused
\end{enumerate}

\subsection{Continuity Planning} \label{sec:continuity}
Once a harm is established, all reasonable efforts to prevent another incident should be taken.
In many cases, this means removing a system from production for a period of time while the remainder of the incident response process is executed.
Some sort of procedure should be established to allow for continuity of operations during this period that is contingent on the severity of the harm.
For example, a self driving car that strikes a pedestrian may require temporarily suspending self-driving across a fleet or limiting where it can be used.
In the case of something like a discriminatory sentencing algorithm, we may simply allow judges to operate as they did before the tool was available, suspending its use. 
Other cases, such as Twitter's image cropping algorithm's ``male gaze'' bias or its aversion to non-Latin text may not rise to the need for continuity response and can remain in production.

In many cases, the scale of these harms -- bad user experience, emotional pain and suffering, loss of life -- can be anticipated, even if the specific harm cannot.
This provides the ability to set up risk-based continuity planning.
Essentially, we seek to answer the question: ``if we have to remove this system from production, what will we do instead?'' to ensure that those who depend in some way on these systems are still able to leverage them, even with limited functionality.

\subsection{Root Cause Analysis}
In security, root cause analysis is used to ask and answer questions about the series of events which led to a security incident, often with a particular focus on the vulnerabilities exploited and why they were not patched.
Even in so-called blameless post-mortems, the root cause analysis seeks to determine what was the weak link in the chain and how said weak link could have been avoided.
In the case of algorithmic harms, a root cause analysis is likely to be much more involved, due to the large number of pieces at play.

The first place to look when a harm occurs is what, if anything, has changed in the system since it most recently functioned at an acceptable level.
If there was an update the morning before an incident, it is prudent to investigate whether or not the previous version of the system would have caused the harm. 
If not, an ablation study should be conducted across the pipeline to identify what components, if any, could be changed to mitigate the harm.
This root cause analysis then informs future penetration tests and threat models to ensure that another incident is not caused by the same cause.

\subsection{Remediating Harms}
After a root cause is identified, it is prudent to remediate both the harms themselves and the causes of said harms.
Remediating a harm depends a lot on the particulars of the harms caused and is currently an issue being openly discussed.
For the teenagers harmed by the promotion of eating disorders on social media~\cite{osullivan2021instagram}, it is unlikely that they will be directly compensated by the organization perpetuating the harm.
Remediating the harm itself is a difficult task that asks much larger questions about who is responsible for these incidents, how the costs are handled, and who, if anyone, owes harmed parties reparations for said harms.
For harms at the scale of COMPAS, the questions grow even larger.
However, as governments like the EU consider revising their product liability regimes to incorporate AI~\cite{eu2021civil}, developers and purveyors of these systems should develop a plan for how to address potential claims against their systems.

Remediating the cause of the harm then, is the more straightforward task -- though by no means is the task simple.
Remediating the harm extends the work of root cause analysis and opens the question of how to fix the root cause. 
In the case of bias, this could be a matter of finding a new dataset, calibrating according to sensitive attributes, leveraging multicalibration~\cite{hebert2018multicalibration}, decision calibration~\cite{zhao2021calibrating}, or some other method.
In other cases, the cause of the harm may necessitate pre- or post-processing of data and decisions to create guardrails.
Yet other cases may require a fundamental reconsideration of the system in use and whether or not it is feasible to have a safe, fair system.
These harms and remediations must be documented to ensure that future projects do not fall into the same trap and can be evaluated using similar methods.

\section{A Responsible AI Development Lifecycle}
\begin{figure*}[ht]
    \centering
    \includegraphics[width=\textwidth]{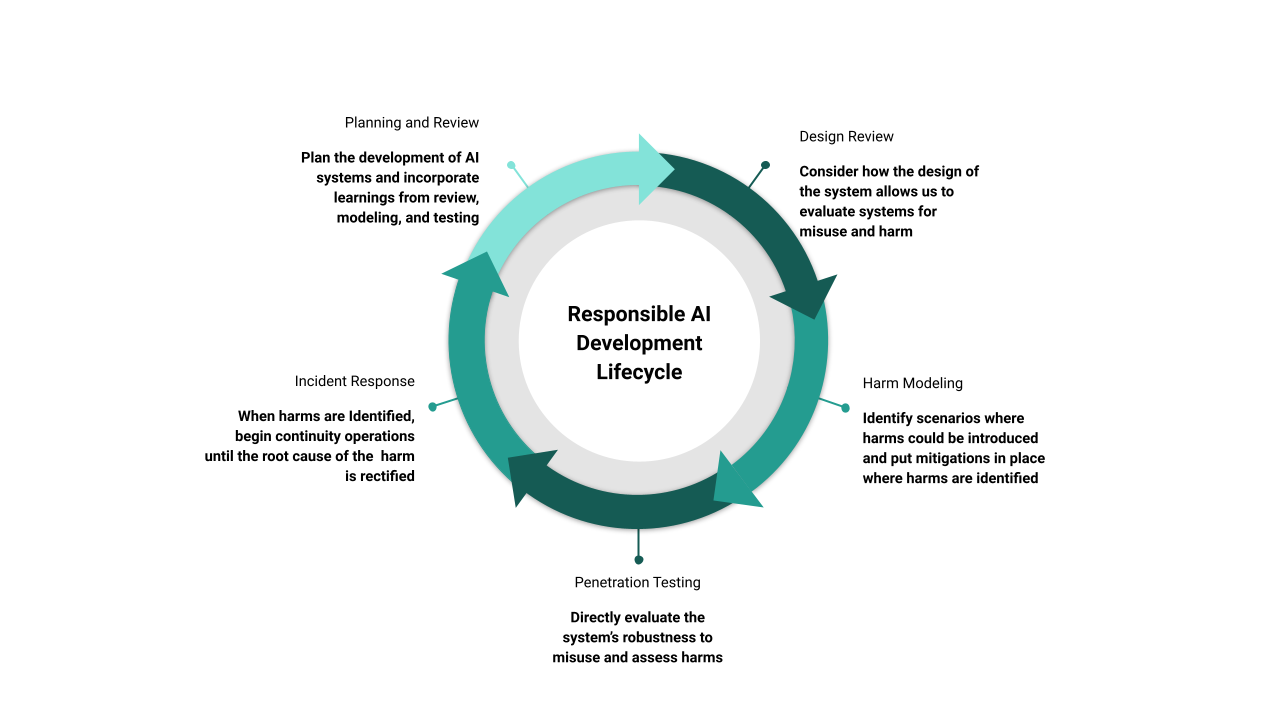}
    \caption{An Illustration of the Responsible AI Development Lifecycle}
    \label{fig:lifecycle}
\end{figure*}

Given the analogies between security and ethical artificial intelligence in Sections~\ref{sec:adapting} and~\ref{sec:incident}, we propose a framework for a responsible AI development lifecycle, illustrated in Figure~\ref{fig:lifecycle}.
Organizations working with artificial intelligence and machine learning have existing processes for design, development, training, testing, and deployment of AI systems.
Though we do not detail those processes here, the proposed framework aims not to replace any part of that process, but rather to augment existing processes with ethical principles.
The responsible AI development lifecycle consists of five steps:
\begin{enumerate}
    \item Planning and Review
    \item Design Review
    \item Harm Modeling
    \item Penetration Testing
    \item Incident Response
\end{enumerate}

Planning and review, as both the first and last step in the process, is intended to revise existing systems and inform the development of new systems.
This step should occur before any new code is written, whether that is in the process of remediating harms or designing a new system.
The planning and review process should look back at prior findings in this system and other systems and set forth the steps that need to be taken in developing this new system.
Additionally, this step is where business continuity planning as outlined in Section~\ref{sec:continuity} should occur.
One critical part of planning and review is the process of documentation -- this involves documenting plans and findings, capturing learnings from incidents, and identifying structures that should be in place ahead of development.

This process leads into the design review step where the overall design of the system is set out.
The intent of the system should be clear, the algorithmic components should be well understood, and data sources should be documented.
In a more mature organization, a design review should include datasheets~\cite{gebru2021datasheets} or model cards~\cite{mitchell2019model} to nail down specific places where there are known issues.
As discussed in Section~\ref{sec:design_review}, the design review should consider not only the artificial intelligence component, but the scaffolding around it and the system as a whole to include logging and auditing, pre-processing, post-processing, and in-processing.

Coming out of the design review step, training and tuning models, deployment, and testing can occur in parallel with harm modeling.
As discussed in Section~\ref{sec:threat_modeling}, this is a good place for counterfactual reasoning -- asking all of the ``what if'' scenarios to understand what can go wrong. 
This process should involve stakeholders from throughout the organization to identify potential harms to users of the system and to external parties.
Where possible, this process should also identify mitigations that can be put in place before the system goes live.
The reason to put mitigations in place ahead of time is twofold: first, having mitigations in place ahead of deployment reduces the likelihood that an individual experiences an identified harm; second, it reduces the cost of putting the mitigation in place since there is no downtime needed to implement it.

When the system is deployed or ready for deployment, we can conduct penetration testing of the system, as in Section~\ref{sec:pen_test}.
This penetration test differs from a traditional penetration test in a number of respects, but is not entirely divorced from the original notion.
Specifically, we are still concerned with discovering security bugs since many of those can be leveraged to cause harm.
Where this approach differs is in taking a broader view of what is in-scope for a penetration test, since we are concerned not only with the possibility of attacks on our models, but also with potential representational harms that are difficult to uncover in security testing.

Assuming that we have established a continuity plan and done our job in the design review phase, there should be good feedback mechanisms for uncovering harms in our systems. 
These can be monitored and spot-corrected over time to ensure that the system is functioning as intended and is not perpetrating harms.
When something goes out of wack or a user or third-party reports a problem, we should initiate our incident response.
As mentioned in Section~\ref{sec:incident}, we should begin by identifying the scope and scale of the harm, then proceed to initiate our business continuity plan if necessary.
Once we have determined the root cause of the harm, we develop a plan to alleviate the problem and proceed back to the top of the process -- reviewing our findings and planning our fixes.

As teams cycle through this process for a particular system, each trip through the cycle should be shorter and easier.
In some cases, the cycle may terminate entirely if the outcome of the review and planning step is a decision not to use an AI system. 
When evaluating the risks and benefits of deploying an AI system, it is important to always consider the reference point of simply not using artificial intelligence 

\section{Conclusion}
This work establishes a framework for developing AI systems responsibly.
We identify two parallel taxonomies of harm: allocative harm versus representational harm and system-on-user harm versus user-on-user harm.
These two taxonomies allow us to develop methods of uncovering, identifying, and classifying harms to users.
Since AI development occurs not in a vacuum but rather as part of a broader development cycle, we view the proposed framework as something that can easily work alongside existing AI system production methods.

The proposed system consists of 5 steps that mirror the standard lifecycle of a software system -- design, development, training, testing, and deployment.
As development proceeds, our framework helps AI developers and stakeholders evaluate their system for potential harms and address them.
This framework differs from existing prescriptions of fairness metrics, explainability methods, and interpretable models due to the shortcomings of those methods computationally, epistemologically, and practically.
Since nearly all systems inevitably change, fail, or prove insufficient, this framework offers an opportunity for iterative ethical improvements without sacrificing practicality. 
We achieve this by analogy with information security, and our framework built on the foundations of the secure software development lifecycle.

This framework makes a number of strong assumptions by necessity.
Specifically, we assume that a ``harm'' can be identified and that given sufficient care, can be resolved.
Crucially, we also assume that organizations actually want to identify and remediate harms to users and are willing to expend effort to improve their systems to that end.
Finally, we assume that the skills to evaluate these models is present in organizations that wish to adopt these principles -- something that we know is untrue for the majority of organizations.

In future work, we aim to address the feasibility of automating parts of this process to reduce the need for specially-skilled individuals.
Today, the creation of documentation around bias, robustness, and other important features of responsible AI is cumbersome and manual.
By offering opportunities to automate and operationalize some of this work, adoption of these processes is eased.

\bibliographystyle{ACM-Reference-Format}
\bibliography{references}


\begin{thebibliography}{73}


\ifx \showCODEN    \undefined \def \showCODEN     #1{\unskip}     \fi
\ifx \showDOI      \undefined \def \showDOI       #1{#1}\fi
\ifx \showISBNx    \undefined \def \showISBNx     #1{\unskip}     \fi
\ifx \showISBNxiii \undefined \def \showISBNxiii  #1{\unskip}     \fi
\ifx \showISSN     \undefined \def \showISSN      #1{\unskip}     \fi
\ifx \showLCCN     \undefined \def \showLCCN      #1{\unskip}     \fi
\ifx \shownote     \undefined \def \shownote      #1{#1}          \fi
\ifx \showarticletitle \undefined \def \showarticletitle #1{#1}   \fi
\ifx \showURL      \undefined \def \showURL       {\relax}        \fi
\providecommand\bibfield[2]{#2}
\providecommand\bibinfo[2]{#2}
\providecommand\natexlab[1]{#1}
\providecommand\showeprint[2][]{arXiv:#2}

\bibitem[Agarwal et~al\mbox{.}(2021)]%
        {agarwal2021towards}
\bibfield{author}{\bibinfo{person}{Sushant Agarwal}, \bibinfo{person}{Shahin
  Jabbari}, \bibinfo{person}{Chirag Agarwal}, \bibinfo{person}{Sohini
  Upadhyay}, \bibinfo{person}{Zhiwei~Steven Wu}, {and}
  \bibinfo{person}{Himabindu Lakkaraju}.} \bibinfo{year}{2021}\natexlab{}.
\newblock \showarticletitle{Towards the Unification and Robustness of
  Perturbation and Gradient Based Explanations}.
\newblock \bibinfo{journal}{\emph{International Conference on Machine
  Learning}}.
\newblock


\bibitem[Ahmed and Kashmoola(2021)]%
        {ahmed2021threats}
\bibfield{author}{\bibinfo{person}{Ibrahim~M Ahmed} {and}
  \bibinfo{person}{Manar~Younis Kashmoola}.} \bibinfo{year}{2021}\natexlab{}.
\newblock \showarticletitle{Threats on Machine Learning Technique by Data
  Poisoning Attack: A Survey}. In \bibinfo{booktitle}{\emph{International
  Conference on Advances in Cyber Security}}. Springer,
  \bibinfo{pages}{586--600}.
\newblock


\bibitem[Al-Rubaie and Chang(2019)]%
        {al2019privacy}
\bibfield{author}{\bibinfo{person}{Mohammad Al-Rubaie} {and}
  \bibinfo{person}{J~Morris Chang}.} \bibinfo{year}{2019}\natexlab{}.
\newblock \showarticletitle{Privacy-preserving machine learning: Threats and
  solutions}.
\newblock \bibinfo{journal}{\emph{IEEE Security \& Privacy}}
  \bibinfo{volume}{17}, \bibinfo{number}{2} (\bibinfo{year}{2019}),
  \bibinfo{pages}{49--58}.
\newblock


\bibitem[Anderson and Roth(2018)]%
        {anderson2018ember}
\bibfield{author}{\bibinfo{person}{Hyrum~S Anderson} {and}
  \bibinfo{person}{Phil Roth}.} \bibinfo{year}{2018}\natexlab{}.
\newblock \showarticletitle{Ember: an open dataset for training static pe
  malware machine learning models}.
\newblock \bibinfo{journal}{\emph{arXiv preprint arXiv:1804.04637}}
  (\bibinfo{year}{2018}).
\newblock


\bibitem[Angwin et~al\mbox{.}(2016)]%
        {angwin2016machine}
\bibfield{author}{\bibinfo{person}{Julia Angwin}, \bibinfo{person}{Jeff
  Larson}, \bibinfo{person}{Surya Mattu}, {and} \bibinfo{person}{Lauren
  Kirchner}.} \bibinfo{year}{2016}\natexlab{}.
\newblock \showarticletitle{Machine bias: There’s software used across the
  country to predict future criminals. And it’s biased against blacks.
  ProPublica (2016)}.
\newblock \bibinfo{journal}{\emph{Google Scholar}} (\bibinfo{year}{2016}),
  \bibinfo{pages}{23}.
\newblock


\bibitem[Babu and Suresh(2012)]%
        {babu2012meta}
\bibfield{author}{\bibinfo{person}{G~Sateesh Babu} {and}
  \bibinfo{person}{Sundaram Suresh}.} \bibinfo{year}{2012}\natexlab{}.
\newblock \showarticletitle{Meta-cognitive neural network for classification
  problems in a sequential learning framework}.
\newblock \bibinfo{journal}{\emph{Neurocomputing}}  \bibinfo{volume}{81}
  (\bibinfo{year}{2012}), \bibinfo{pages}{86--96}.
\newblock


\bibitem[Bansal et~al\mbox{.}(2021)]%
        {bansal2021does}
\bibfield{author}{\bibinfo{person}{Gagan Bansal}, \bibinfo{person}{Tongshuang
  Wu}, \bibinfo{person}{Joyce Zhou}, \bibinfo{person}{Raymond Fok},
  \bibinfo{person}{Besmira Nushi}, \bibinfo{person}{Ece Kamar},
  \bibinfo{person}{Marco~Tulio Ribeiro}, {and} \bibinfo{person}{Daniel Weld}.}
  \bibinfo{year}{2021}\natexlab{}.
\newblock \showarticletitle{Does the whole exceed its parts? the effect of ai
  explanations on complementary team performance}. In
  \bibinfo{booktitle}{\emph{Proceedings of the 2021 CHI Conference on Human
  Factors in Computing Systems}}. \bibinfo{pages}{1--16}.
\newblock


\bibitem[Barocas et~al\mbox{.}(2017)]%
        {barocas2017fairness}
\bibfield{author}{\bibinfo{person}{Solon Barocas}, \bibinfo{person}{Moritz
  Hardt}, {and} \bibinfo{person}{Arvind Narayanan}.}
  \bibinfo{year}{2017}\natexlab{}.
\newblock \showarticletitle{Fairness in machine learning}.
\newblock \bibinfo{journal}{\emph{Nips tutorial}}  \bibinfo{volume}{1}
  (\bibinfo{year}{2017}), \bibinfo{pages}{2017}.
\newblock


\bibitem[Birhane et~al\mbox{.}(2022)]%
        {birhane2022auditing}
\bibfield{author}{\bibinfo{person}{Abeba Birhane}, \bibinfo{person}{Vinay~Uday
  Prabhu}, {and} \bibinfo{person}{John Whaley}.}
  \bibinfo{year}{2022}\natexlab{}.
\newblock \showarticletitle{Auditing Saliency Cropping Algorithms}. In
  \bibinfo{booktitle}{\emph{Proceedings of the IEEE/CVF Winter Conference on
  Applications of Computer Vision}}. \bibinfo{pages}{4051--4059}.
\newblock


\bibitem[Boucher et~al\mbox{.}(2022)]%
        {boucher2022bad}
\bibfield{author}{\bibinfo{person}{Nicholas Boucher}, \bibinfo{person}{Ilia
  Shumailov}, \bibinfo{person}{Ross Anderson}, {and} \bibinfo{person}{Nicolas
  Papernot}.} \bibinfo{year}{2022}\natexlab{}.
\newblock \showarticletitle{Bad characters: Imperceptible nlp attacks}.
\newblock \bibinfo{journal}{\emph{43rd IEEE Symposium on Security and Privacy}}
  (\bibinfo{year}{2022}).
\newblock


\bibitem[Carlini et~al\mbox{.}(2021)]%
        {carlini2021extracting}
\bibfield{author}{\bibinfo{person}{Nicholas Carlini}, \bibinfo{person}{Florian
  Tramer}, \bibinfo{person}{Eric Wallace}, \bibinfo{person}{Matthew Jagielski},
  \bibinfo{person}{Ariel Herbert-Voss}, \bibinfo{person}{Katherine Lee},
  \bibinfo{person}{Adam Roberts}, \bibinfo{person}{Tom Brown},
  \bibinfo{person}{Dawn Song}, \bibinfo{person}{Ulfar Erlingsson},
  {et~al\mbox{.}}} \bibinfo{year}{2021}\natexlab{}.
\newblock \showarticletitle{Extracting training data from large language
  models}. In \bibinfo{booktitle}{\emph{30th USENIX Security Symposium (USENIX
  Security 21)}}. \bibinfo{pages}{2633--2650}.
\newblock


\bibitem[Carvalho et~al\mbox{.}(2019)]%
        {carvalho2019machine}
\bibfield{author}{\bibinfo{person}{Diogo~V Carvalho},
  \bibinfo{person}{Eduardo~M Pereira}, {and} \bibinfo{person}{Jaime~S
  Cardoso}.} \bibinfo{year}{2019}\natexlab{}.
\newblock \showarticletitle{Machine learning interpretability: A survey on
  methods and metrics}.
\newblock \bibinfo{journal}{\emph{Electronics}} \bibinfo{volume}{8},
  \bibinfo{number}{8} (\bibinfo{year}{2019}), \bibinfo{pages}{832}.
\newblock


\bibitem[Chen and Deng(2006)]%
        {chen2006settling}
\bibfield{author}{\bibinfo{person}{Xi Chen} {and} \bibinfo{person}{Xiaotie
  Deng}.} \bibinfo{year}{2006}\natexlab{}.
\newblock \showarticletitle{Settling the complexity of two-player Nash
  equilibrium}. In \bibinfo{booktitle}{\emph{2006 47th Annual IEEE Symposium on
  Foundations of Computer Science (FOCS'06)}}. IEEE, \bibinfo{pages}{261--272}.
\newblock


\bibitem[Chess and McGraw(2004)]%
        {chess2004static}
\bibfield{author}{\bibinfo{person}{Brian Chess} {and} \bibinfo{person}{Gary
  McGraw}.} \bibinfo{year}{2004}\natexlab{}.
\newblock \showarticletitle{Static analysis for security}.
\newblock \bibinfo{journal}{\emph{IEEE security \& privacy}}
  \bibinfo{volume}{2}, \bibinfo{number}{6} (\bibinfo{year}{2004}),
  \bibinfo{pages}{76--79}.
\newblock


\bibitem[Choquette-Choo et~al\mbox{.}(2021)]%
        {choquette2021label}
\bibfield{author}{\bibinfo{person}{Christopher~A Choquette-Choo},
  \bibinfo{person}{Florian Tramer}, \bibinfo{person}{Nicholas Carlini}, {and}
  \bibinfo{person}{Nicolas Papernot}.} \bibinfo{year}{2021}\natexlab{}.
\newblock \showarticletitle{Label-only membership inference attacks}. In
  \bibinfo{booktitle}{\emph{International Conference on Machine Learning}}.
  PMLR, \bibinfo{pages}{1964--1974}.
\newblock


\bibitem[Chowdhury and Williams(2021)]%
        {chowdhury2021introducing}
\bibfield{author}{\bibinfo{person}{Rumman Chowdhury} {and}
  \bibinfo{person}{Jutta Williams}.} \bibinfo{year}{2021}\natexlab{}.
\newblock \bibinfo{title}{Introducing Twitter's first algorithmic bias bounty
  challenge}.
\newblock
  \bibinfo{howpublished}{\url{https://blog.twitter.com/engineering/en_us/topics/insights/2021/algorithmic-bias-bounty-challenge}}.
\newblock
\newblock
\shownote{[Online; accessed 17-January-2022]}.


\bibitem[Cohen(1987)]%
        {cohen1987computer}
\bibfield{author}{\bibinfo{person}{Fred Cohen}.}
  \bibinfo{year}{1987}\natexlab{}.
\newblock \showarticletitle{Computer viruses: theory and experiments}.
\newblock \bibinfo{journal}{\emph{Computers \& security}} \bibinfo{volume}{6},
  \bibinfo{number}{1} (\bibinfo{year}{1987}), \bibinfo{pages}{22--35}.
\newblock


\bibitem[Commission et~al\mbox{.}(2021a)]%
        {eu2021civil}
\bibfield{author}{\bibinfo{person}{EU Commission} {et~al\mbox{.}}}
  \bibinfo{year}{2021}\natexlab{a}.
\newblock \showarticletitle{Civil liability – adapting liability rules to the
  digital age and artificial intelligence}.
\newblock
  \bibinfo{howpublished}{\url{https://ec.europa.eu/info/law/better-regulation/have-your-say/initiatives/12979-Civil-liability-adapting-liability-rules-to-the-digital-age-and-artificial-intelligence/public-consultation_en}}.
\newblock  (\bibinfo{year}{2021}).
\newblock
\newblock
\shownote{[Online; accessed 3-February-2022]}.


\bibitem[Commission et~al\mbox{.}(2021b)]%
        {eu2021proposal}
\bibfield{author}{\bibinfo{person}{EU Commission} {et~al\mbox{.}}}
  \bibinfo{year}{2021}\natexlab{b}.
\newblock \showarticletitle{Proposal for a regulation of the European
  Parliament and of the Council laying down harmonised rules on artificial
  intelligence (Artificial Intelligence Act) and amending certain Union
  legislative acts}.
\newblock \bibinfo{journal}{\emph{COM (2021)}}  \bibinfo{volume}{206}
  (\bibinfo{year}{2021}).
\newblock


\bibitem[Crawford(2017)]%
        {crawford2017trouble}
\bibfield{author}{\bibinfo{person}{Kate Crawford}.}
  \bibinfo{year}{2017}\natexlab{}.
\newblock \bibinfo{title}{The Trouble With Bias}.
\newblock
\newblock
\urldef\tempurl%
\url{https://nips.cc/Conferences/2017/Schedule?showEvent=8742}
\showURL{%
\tempurl}
\newblock
\shownote{Thirty-first Conference on Neural Information Processing Systems
  Keynote Presentation}.


\bibitem[Dai et~al\mbox{.}(2021)]%
        {dai2021fair}
\bibfield{author}{\bibinfo{person}{Jessica Dai}, \bibinfo{person}{Sina
  Fazelpour}, {and} \bibinfo{person}{Zachary Lipton}.}
  \bibinfo{year}{2021}\natexlab{}.
\newblock \showarticletitle{Fair machine learning under partial compliance}. In
  \bibinfo{booktitle}{\emph{Proceedings of the 2021 AAAI/ACM Conference on AI,
  Ethics, and Society}}. \bibinfo{pages}{55--65}.
\newblock


\bibitem[Dastin(2018)]%
        {dastin2018amazon}
\bibfield{author}{\bibinfo{person}{Jeffrey Dastin}.}
  \bibinfo{year}{2018}\natexlab{}.
\newblock \bibinfo{title}{Amazon scraps secret AI recruiting tool that showed
  bias against women}.
\newblock
\newblock


\bibitem[Dowd et~al\mbox{.}(2006)]%
        {dowd2006art}
\bibfield{author}{\bibinfo{person}{Mark Dowd}, \bibinfo{person}{John McDonald},
  {and} \bibinfo{person}{Justin Schuh}.} \bibinfo{year}{2006}\natexlab{}.
\newblock \bibinfo{booktitle}{\emph{The art of software security assessment:
  Identifying and preventing software vulnerabilities}}.
\newblock \bibinfo{publisher}{Pearson Education}.
\newblock


\bibitem[Dwork et~al\mbox{.}(2012)]%
        {dwork2012fairness}
\bibfield{author}{\bibinfo{person}{Cynthia Dwork}, \bibinfo{person}{Moritz
  Hardt}, \bibinfo{person}{Toniann Pitassi}, \bibinfo{person}{Omer Reingold},
  {and} \bibinfo{person}{Richard Zemel}.} \bibinfo{year}{2012}\natexlab{}.
\newblock \showarticletitle{Fairness through awareness}. In
  \bibinfo{booktitle}{\emph{Proceedings of the 3rd innovations in theoretical
  computer science conference}}. \bibinfo{pages}{214--226}.
\newblock


\bibitem[Fazelpour and Lipton(2020)]%
        {fazelpour2020algorithmic}
\bibfield{author}{\bibinfo{person}{Sina Fazelpour} {and}
  \bibinfo{person}{Zachary~C Lipton}.} \bibinfo{year}{2020}\natexlab{}.
\newblock \showarticletitle{Algorithmic fairness from a non-ideal perspective}.
  In \bibinfo{booktitle}{\emph{Proceedings of the AAAI/ACM Conference on AI,
  Ethics, and Society}}. \bibinfo{pages}{57--63}.
\newblock


\bibitem[Fingas(2021)]%
        {fingas2021twitter}
\bibfield{author}{\bibinfo{person}{Jon Fingas}.}
  \bibinfo{year}{2021}\natexlab{}.
\newblock \bibinfo{title}{Twitter's AI bounty program reveals bias toward
  young, pretty white people}.
\newblock
  \bibinfo{howpublished}{\url{https://www.engadget.com/twitter-ai-bias-beauty-filters-133210055.html}}.
\newblock
\newblock
\shownote{[Online; accessed 19-January-2022]}.


\bibitem[Fredrikson et~al\mbox{.}(2015)]%
        {fredrikson2015model}
\bibfield{author}{\bibinfo{person}{Matt Fredrikson}, \bibinfo{person}{Somesh
  Jha}, {and} \bibinfo{person}{Thomas Ristenpart}.}
  \bibinfo{year}{2015}\natexlab{}.
\newblock \showarticletitle{Model inversion attacks that exploit confidence
  information and basic countermeasures}. In
  \bibinfo{booktitle}{\emph{Proceedings of the 22nd ACM SIGSAC conference on
  computer and communications security}}. \bibinfo{pages}{1322--1333}.
\newblock


\bibitem[Gebru et~al\mbox{.}(2021)]%
        {gebru2021datasheets}
\bibfield{author}{\bibinfo{person}{Timnit Gebru}, \bibinfo{person}{Jamie
  Morgenstern}, \bibinfo{person}{Briana Vecchione},
  \bibinfo{person}{Jennifer~Wortman Vaughan}, \bibinfo{person}{Hanna Wallach},
  \bibinfo{person}{Hal~Daum{\'e} Iii}, {and} \bibinfo{person}{Kate Crawford}.}
  \bibinfo{year}{2021}\natexlab{}.
\newblock \showarticletitle{Datasheets for datasets}.
\newblock \bibinfo{journal}{\emph{Commun. ACM}} \bibinfo{volume}{64},
  \bibinfo{number}{12} (\bibinfo{year}{2021}), \bibinfo{pages}{86--92}.
\newblock


\bibitem[H{\'e}bert-Johnson et~al\mbox{.}(2018)]%
        {hebert2018multicalibration}
\bibfield{author}{\bibinfo{person}{Ursula H{\'e}bert-Johnson},
  \bibinfo{person}{Michael Kim}, \bibinfo{person}{Omer Reingold}, {and}
  \bibinfo{person}{Guy Rothblum}.} \bibinfo{year}{2018}\natexlab{}.
\newblock \showarticletitle{Multicalibration: Calibration for the
  (computationally-identifiable) masses}. In
  \bibinfo{booktitle}{\emph{International Conference on Machine Learning}}.
  PMLR, \bibinfo{pages}{1939--1948}.
\newblock


\bibitem[Howard and Lipner(2006)]%
        {howard2006security}
\bibfield{author}{\bibinfo{person}{Michael Howard} {and} \bibinfo{person}{Steve
  Lipner}.} \bibinfo{year}{2006}\natexlab{}.
\newblock \bibinfo{booktitle}{\emph{The security development lifecycle}}.
  Vol.~\bibinfo{volume}{8}.
\newblock \bibinfo{publisher}{Microsoft Press Redmond}.
\newblock


\bibitem[Hussain(2020)]%
        {hussain2020privacyraven}
\bibfield{author}{\bibinfo{person}{Suha Hussain}.}
  \bibinfo{year}{2020}\natexlab{}.
\newblock \bibinfo{title}{PrivacyRaven Has Left the Nest}.
\newblock
  \bibinfo{howpublished}{\url{https://blog.trailofbits.com/2020/10/08/privacyraven-has-left-the-nest/}}.
\newblock
\newblock
\shownote{[Online; accessed 19-January-2022]}.


\bibitem[Kaur et~al\mbox{.}(2020)]%
        {kaur2020interpreting}
\bibfield{author}{\bibinfo{person}{Harmanpreet Kaur}, \bibinfo{person}{Harsha
  Nori}, \bibinfo{person}{Samuel Jenkins}, \bibinfo{person}{Rich Caruana},
  \bibinfo{person}{Hanna Wallach}, {and} \bibinfo{person}{Jennifer
  Wortman~Vaughan}.} \bibinfo{year}{2020}\natexlab{}.
\newblock \showarticletitle{Interpreting interpretability: understanding data
  scientists' use of interpretability tools for machine learning}. In
  \bibinfo{booktitle}{\emph{Proceedings of the 2020 CHI conference on human
  factors in computing systems}}. \bibinfo{pages}{1--14}.
\newblock


\bibitem[Kleinberg et~al\mbox{.}(2017)]%
        {kleinberg2017inherent}
\bibfield{author}{\bibinfo{person}{Jon Kleinberg}, \bibinfo{person}{Sendhil
  Mullainathan}, {and} \bibinfo{person}{Manish Raghavan}.}
  \bibinfo{year}{2017}\natexlab{}.
\newblock \showarticletitle{Inherent Trade-Offs in the Fair Determination of
  Risk Scores}. In \bibinfo{booktitle}{\emph{8th Innovations in Theoretical
  Computer Science Conference (ITCS 2017)}}. Schloss Dagstuhl-Leibniz-Zentrum
  fuer Informatik.
\newblock


\bibitem[Kloft and Laskov(2007)]%
        {kloft2007poisoning}
\bibfield{author}{\bibinfo{person}{Marius Kloft} {and} \bibinfo{person}{Pavel
  Laskov}.} \bibinfo{year}{2007}\natexlab{}.
\newblock \showarticletitle{A poisoning attack against online anomaly
  detection}. In \bibinfo{booktitle}{\emph{NIPS Workshop on Machine Learning in
  Adversarial Environments for Computer Security}}, Vol.~\bibinfo{volume}{19}.
  Citeseer.
\newblock


\bibitem[Kr{\"o}ger et~al\mbox{.}(2021)]%
        {kroger2021data}
\bibfield{author}{\bibinfo{person}{Jacob~Leon Kr{\"o}ger},
  \bibinfo{person}{Milagros Miceli}, {and} \bibinfo{person}{Florian
  M{\"u}ller}.} \bibinfo{year}{2021}\natexlab{}.
\newblock \showarticletitle{How Data Can Be Used Against People: A
  Classification of Personal Data Misuses}.
\newblock \bibinfo{journal}{\emph{Available at SSRN 3887097}}
  (\bibinfo{year}{2021}).
\newblock


\bibitem[Kumar et~al\mbox{.}(2020b)]%
        {kumar2020problems}
\bibfield{author}{\bibinfo{person}{I~Elizabeth Kumar}, \bibinfo{person}{Suresh
  Venkatasubramanian}, \bibinfo{person}{Carlos Scheidegger}, {and}
  \bibinfo{person}{Sorelle Friedler}.} \bibinfo{year}{2020}\natexlab{b}.
\newblock \showarticletitle{Problems with Shapley-value-based explanations as
  feature importance measures}. In \bibinfo{booktitle}{\emph{International
  Conference on Machine Learning}}. PMLR, \bibinfo{pages}{5491--5500}.
\newblock


\bibitem[Kumar et~al\mbox{.}(2020a)]%
        {kumar2020adversarial}
\bibfield{author}{\bibinfo{person}{Ram Shankar~Siva Kumar},
  \bibinfo{person}{Magnus Nystr{\"o}m}, \bibinfo{person}{John Lambert},
  \bibinfo{person}{Andrew Marshall}, \bibinfo{person}{Mario Goertzel},
  \bibinfo{person}{Andi Comissoneru}, \bibinfo{person}{Matt Swann}, {and}
  \bibinfo{person}{Sharon Xia}.} \bibinfo{year}{2020}\natexlab{a}.
\newblock \showarticletitle{Adversarial machine learning-industry
  perspectives}. In \bibinfo{booktitle}{\emph{2020 IEEE Security and Privacy
  Workshops (SPW)}}. IEEE, \bibinfo{pages}{69--75}.
\newblock


\bibitem[Kumar et~al\mbox{.}(2021)]%
        {kumar2021meta}
\bibfield{author}{\bibinfo{person}{Sannidhi~P Kumar}, \bibinfo{person}{Chandan
  Gautam}, {and} \bibinfo{person}{Suresh Sundaram}.}
  \bibinfo{year}{2021}\natexlab{}.
\newblock \showarticletitle{Meta-Cognition-Based Simple And Effective Approach
  To Object Detection}. In \bibinfo{booktitle}{\emph{ICASSP 2021-2021 IEEE
  International Conference on Acoustics, Speech and Signal Processing
  (ICASSP)}}. IEEE, \bibinfo{pages}{3795--3799}.
\newblock


\bibitem[Lipton et~al\mbox{.}(2004)]%
        {lipton2004approximately}
\bibfield{author}{\bibinfo{person}{Richard~J Lipton},
  \bibinfo{person}{Evangelos Markakis}, \bibinfo{person}{Elchanan Mossel},
  {and} \bibinfo{person}{Amin Saberi}.} \bibinfo{year}{2004}\natexlab{}.
\newblock \showarticletitle{On approximately fair allocations of indivisible
  goods}. In \bibinfo{booktitle}{\emph{Proceedings of the 5th ACM Conference on
  Electronic Commerce}}. \bibinfo{pages}{125--131}.
\newblock


\bibitem[Lipton(2018)]%
        {lipton2018mythos}
\bibfield{author}{\bibinfo{person}{Zachary~C Lipton}.}
  \bibinfo{year}{2018}\natexlab{}.
\newblock \showarticletitle{The Mythos of Model Interpretability: In machine
  learning, the concept of interpretability is both important and slippery.}
\newblock \bibinfo{journal}{\emph{Queue}} \bibinfo{volume}{16},
  \bibinfo{number}{3} (\bibinfo{year}{2018}), \bibinfo{pages}{31--57}.
\newblock


\bibitem[Lundberg and Lee(2017)]%
        {lundberg2017unified}
\bibfield{author}{\bibinfo{person}{Scott~M Lundberg} {and}
  \bibinfo{person}{Su-In Lee}.} \bibinfo{year}{2017}\natexlab{}.
\newblock \showarticletitle{A unified approach to interpreting model
  predictions}. In \bibinfo{booktitle}{\emph{Proceedings of the 31st
  international conference on neural information processing systems}}.
  \bibinfo{pages}{4768--4777}.
\newblock


\bibitem[Marshall(1999)]%
        {marshall1999metacat}
\bibfield{author}{\bibinfo{person}{James~B Marshall}.}
  \bibinfo{year}{1999}\natexlab{}.
\newblock \bibinfo{booktitle}{\emph{Metacat: A self-watching cognitive
  architecture for analogy-making and high-level perception}}.
\newblock \bibinfo{publisher}{Indiana University}.
\newblock


\bibitem[Mitchell et~al\mbox{.}(2019)]%
        {mitchell2019model}
\bibfield{author}{\bibinfo{person}{Margaret Mitchell}, \bibinfo{person}{Simone
  Wu}, \bibinfo{person}{Andrew Zaldivar}, \bibinfo{person}{Parker Barnes},
  \bibinfo{person}{Lucy Vasserman}, \bibinfo{person}{Ben Hutchinson},
  \bibinfo{person}{Elena Spitzer}, \bibinfo{person}{Inioluwa~Deborah Raji},
  {and} \bibinfo{person}{Timnit Gebru}.} \bibinfo{year}{2019}\natexlab{}.
\newblock \showarticletitle{Model cards for model reporting}. In
  \bibinfo{booktitle}{\emph{Proceedings of the conference on fairness,
  accountability, and transparency}}. \bibinfo{pages}{220--229}.
\newblock


\bibitem[Molnar(2020)]%
        {molnar2020interpretable}
\bibfield{author}{\bibinfo{person}{Christoph Molnar}.}
  \bibinfo{year}{2020}\natexlab{}.
\newblock \bibinfo{booktitle}{\emph{Interpretable machine learning}}.
\newblock \bibinfo{publisher}{Lulu. com}.
\newblock


\bibitem[Moody et~al\mbox{.}(2018)]%
        {moody2018toward}
\bibfield{author}{\bibinfo{person}{Gregory~D Moody}, \bibinfo{person}{Mikko
  Siponen}, {and} \bibinfo{person}{Seppo Pahnila}.}
  \bibinfo{year}{2018}\natexlab{}.
\newblock \showarticletitle{Toward a unified model of information security
  policy compliance.}
\newblock \bibinfo{journal}{\emph{MIS quarterly}} \bibinfo{volume}{42},
  \bibinfo{number}{1} (\bibinfo{year}{2018}).
\newblock


\bibitem[of~Standards and Technology(2021)]%
        {nist2021rmf}
\bibfield{author}{\bibinfo{person}{National~Institute of Standards} {and}
  \bibinfo{person}{Technology}.} \bibinfo{year}{2021}\natexlab{}.
\newblock \bibinfo{title}{AI Risk Management Framework Concept Paper}.
\newblock
  \bibinfo{howpublished}{\url{https://www.nist.gov/itl/ai-risk-management-framework}}.
\newblock
\newblock
\shownote{[Online; accessed 10-January-2022]}.


\bibitem[of~State~Legislatures(2021)]%
        {ncsl2021legislation}
\bibfield{author}{\bibinfo{person}{National~Conference of State~Legislatures}.}
  \bibinfo{year}{2021}\natexlab{}.
\newblock \bibinfo{title}{Legislation Related to Artificial Intelligence}.
\newblock
  \bibinfo{howpublished}{\url{https://www.ncsl.org/research/telecommunications-and-information-technology/2020-legislation-related-to-artificial-intelligence.aspx}}.
\newblock
\newblock
\shownote{[Online; accessed 10-January-2022]}.


\bibitem[O'Sullivan et~al\mbox{.}(2021)]%
        {osullivan2021instagram}
\bibfield{author}{\bibinfo{person}{Donie O'Sullivan}, \bibinfo{person}{Clare
  Duffy}, {and} \bibinfo{person}{Sarah Jorgensen}.}
  \bibinfo{year}{2021}\natexlab{}.
\newblock \bibinfo{title}{Instagram promoted pages glorifying eating disorders
  to teen accounts}.
\newblock
  \bibinfo{howpublished}{\url{https://www.cnn.com/2021/10/04/tech/instagram-facebook-eating-disorders/index.html}}.
\newblock
\newblock
\shownote{[Online; accessed 27-January-2022]}.


\bibitem[Pearce and Kumar(2021)]%
        {pearce2021ai}
\bibfield{author}{\bibinfo{person}{Will Pearce} {and} \bibinfo{person}{Ram
  Shankar~Siva Kumar}.} \bibinfo{year}{2021}\natexlab{}.
\newblock \bibinfo{title}{AI Security Risk Assessment Using Counterfit}.
\newblock
  \bibinfo{howpublished}{\url{https://www.microsoft.com/security/blog/2021/05/03/ai-security-risk-assessment-using-counterfit/}}.
\newblock
\newblock
\shownote{[Online; accessed 19-January-2022]}.


\bibitem[Pearl and Mackenzie(2018)]%
        {pearl2018book}
\bibfield{author}{\bibinfo{person}{Judea Pearl} {and} \bibinfo{person}{Dana
  Mackenzie}.} \bibinfo{year}{2018}\natexlab{}.
\newblock \bibinfo{booktitle}{\emph{The book of why: the new science of cause
  and effect}}.
\newblock \bibinfo{publisher}{Basic books}.
\newblock


\bibitem[Pitropakis et~al\mbox{.}(2019)]%
        {pitropakis2019taxonomy}
\bibfield{author}{\bibinfo{person}{Nikolaos Pitropakis},
  \bibinfo{person}{Emmanouil Panaousis}, \bibinfo{person}{Thanassis
  Giannetsos}, \bibinfo{person}{Eleftherios Anastasiadis}, {and}
  \bibinfo{person}{George Loukas}.} \bibinfo{year}{2019}\natexlab{}.
\newblock \showarticletitle{A taxonomy and survey of attacks against machine
  learning}.
\newblock \bibinfo{journal}{\emph{Computer Science Review}}
  \bibinfo{volume}{34} (\bibinfo{year}{2019}), \bibinfo{pages}{100199}.
\newblock


\bibitem[Povolny and Trivedi(2020)]%
        {povolny2020model}
\bibfield{author}{\bibinfo{person}{Steve Povolny} {and}
  \bibinfo{person}{Shivangee Trivedi}.} \bibinfo{year}{2020}\natexlab{}.
\newblock \showarticletitle{Model hacking ADAS to pave safer roads for
  autonomous vehicles}.
\newblock \bibinfo{journal}{\emph{McAfee Advanced Threat Research}}
  (\bibinfo{year}{2020}).
\newblock


\bibitem[Ribeiro et~al\mbox{.}(2016)]%
        {ribeiro2016should}
\bibfield{author}{\bibinfo{person}{Marco~Tulio Ribeiro},
  \bibinfo{person}{Sameer Singh}, {and} \bibinfo{person}{Carlos Guestrin}.}
  \bibinfo{year}{2016}\natexlab{}.
\newblock \showarticletitle{" Why should i trust you?" Explaining the
  predictions of any classifier}. In \bibinfo{booktitle}{\emph{Proceedings of
  the 22nd ACM SIGKDD international conference on knowledge discovery and data
  mining}}. \bibinfo{pages}{1135--1144}.
\newblock


\bibitem[Ringer et~al\mbox{.}(2020)]%
        {ringer2020qed}
\bibfield{author}{\bibinfo{person}{Talia Ringer}, \bibinfo{person}{Karl
  Palmskog}, \bibinfo{person}{Ilya Sergey}, \bibinfo{person}{Milos Gligoric},
  {and} \bibinfo{person}{Zachary Tatlock}.} \bibinfo{year}{2020}\natexlab{}.
\newblock \showarticletitle{QED at large: A survey of engineering of formally
  verified software}.
\newblock \bibinfo{journal}{\emph{arXiv preprint arXiv:2003.06458}}
  (\bibinfo{year}{2020}).
\newblock


\bibitem[Robertson and Webb(1998)]%
        {robertson1998cake}
\bibfield{author}{\bibinfo{person}{Jack Robertson} {and}
  \bibinfo{person}{William Webb}.} \bibinfo{year}{1998}\natexlab{}.
\newblock \bibinfo{booktitle}{\emph{Cake-cutting algorithms: Be fair if you
  can}}.
\newblock \bibinfo{publisher}{CRC Press}.
\newblock


\bibitem[Roy(2020)]%
        {roy2020high}
\bibfield{author}{\bibinfo{person}{Prameet~P Roy}.}
  \bibinfo{year}{2020}\natexlab{}.
\newblock \showarticletitle{A High-Level Comparison between the NIST Cyber
  Security Framework and the ISO 27001 Information Security Standard}. In
  \bibinfo{booktitle}{\emph{2020 National Conference on Emerging Trends on
  Sustainable Technology and Engineering Applications (NCETSTEA)}}. IEEE,
  \bibinfo{pages}{1--3}.
\newblock


\bibitem[Schmidt and Biessmann(2019)]%
        {schmidt2019quantifying}
\bibfield{author}{\bibinfo{person}{Philipp Schmidt} {and}
  \bibinfo{person}{Felix Biessmann}.} \bibinfo{year}{2019}\natexlab{}.
\newblock \showarticletitle{Quantifying interpretability and trust in machine
  learning systems}.
\newblock \bibinfo{journal}{\emph{AAAI-19 Workshop on Network Interpretability
  for Deep learning}} (\bibinfo{year}{2019}).
\newblock


\bibitem[Selvaraju et~al\mbox{.}(2017)]%
        {selvaraju2017grad}
\bibfield{author}{\bibinfo{person}{Ramprasaath~R Selvaraju},
  \bibinfo{person}{Michael Cogswell}, \bibinfo{person}{Abhishek Das},
  \bibinfo{person}{Ramakrishna Vedantam}, \bibinfo{person}{Devi Parikh}, {and}
  \bibinfo{person}{Dhruv Batra}.} \bibinfo{year}{2017}\natexlab{}.
\newblock \showarticletitle{Grad-cam: Visual explanations from deep networks
  via gradient-based localization}. In \bibinfo{booktitle}{\emph{Proceedings of
  the IEEE international conference on computer vision}}.
  \bibinfo{pages}{618--626}.
\newblock


\bibitem[Shepardson and Jin(2021)]%
        {shepardson2021us}
\bibfield{author}{\bibinfo{person}{David Shepardson} {and}
  \bibinfo{person}{Hyunjoo Jin}.} \bibinfo{year}{2021}\natexlab{}.
\newblock \showarticletitle{U.S. probing fatal Tesla crash that killed
  pedestrian}.
\newblock
  \bibinfo{howpublished}{\url{https://www.reuters.com/business/autos-transportation/us-probing-fatal-tesla-crash-that-killed-pedestrian-2021-09-03/}}.
\newblock \bibinfo{journal}{\emph{Reuters}} (\bibinfo{year}{2021}).
\newblock
\newblock
\shownote{[Online; accessed 17-January-2022]}.


\bibitem[Shokri et~al\mbox{.}(2017)]%
        {shokri2017membership}
\bibfield{author}{\bibinfo{person}{Reza Shokri}, \bibinfo{person}{Marco
  Stronati}, \bibinfo{person}{Congzheng Song}, {and} \bibinfo{person}{Vitaly
  Shmatikov}.} \bibinfo{year}{2017}\natexlab{}.
\newblock \showarticletitle{Membership inference attacks against machine
  learning models}. In \bibinfo{booktitle}{\emph{2017 IEEE symposium on
  security and privacy (SP)}}. IEEE, \bibinfo{pages}{3--18}.
\newblock


\bibitem[Shostack(2014)]%
        {shostack2014threat}
\bibfield{author}{\bibinfo{person}{Adam Shostack}.}
  \bibinfo{year}{2014}\natexlab{}.
\newblock \bibinfo{booktitle}{\emph{Threat modeling: Designing for security}}.
\newblock \bibinfo{publisher}{John Wiley \& Sons}.
\newblock


\bibitem[Shumailov et~al\mbox{.}(2021)]%
        {shumailov2021sponge}
\bibfield{author}{\bibinfo{person}{Ilia Shumailov}, \bibinfo{person}{Yiren
  Zhao}, \bibinfo{person}{Daniel Bates}, \bibinfo{person}{Nicolas Papernot},
  \bibinfo{person}{Robert Mullins}, {and} \bibinfo{person}{Ross Anderson}.}
  \bibinfo{year}{2021}\natexlab{}.
\newblock \showarticletitle{Sponge examples: Energy-latency attacks on neural
  networks}. In \bibinfo{booktitle}{\emph{2021 IEEE European Symposium on
  Security and Privacy (EuroS\&P)}}. IEEE, \bibinfo{pages}{212--231}.
\newblock


\bibitem[Slack et~al\mbox{.}(2020)]%
        {slack2020fooling}
\bibfield{author}{\bibinfo{person}{Dylan Slack}, \bibinfo{person}{Sophie
  Hilgard}, \bibinfo{person}{Emily Jia}, \bibinfo{person}{Sameer Singh}, {and}
  \bibinfo{person}{Himabindu Lakkaraju}.} \bibinfo{year}{2020}\natexlab{}.
\newblock \showarticletitle{Fooling lime and shap: Adversarial attacks on post
  hoc explanation methods}. In \bibinfo{booktitle}{\emph{Proceedings of the
  AAAI/ACM Conference on AI, Ethics, and Society}}. \bibinfo{pages}{180--186}.
\newblock


\bibitem[Steel et~al\mbox{.}(2021)]%
        {steel2021information}
\bibfield{author}{\bibinfo{person}{Daniel Steel}, \bibinfo{person}{Sina
  Fazelpour}, \bibinfo{person}{Bianca Crewe}, {and} \bibinfo{person}{Kinley
  Gillette}.} \bibinfo{year}{2021}\natexlab{}.
\newblock \showarticletitle{Information elaboration and epistemic effects of
  diversity}.
\newblock \bibinfo{journal}{\emph{Synthese}} \bibinfo{volume}{198},
  \bibinfo{number}{2} (\bibinfo{year}{2021}), \bibinfo{pages}{1287--1307}.
\newblock


\bibitem[Sulistyowati et~al\mbox{.}(2020)]%
        {sulistyowati2020comparative}
\bibfield{author}{\bibinfo{person}{Diah Sulistyowati}, \bibinfo{person}{Fitri
  Handayani}, {and} \bibinfo{person}{Yohan Suryanto}.}
  \bibinfo{year}{2020}\natexlab{}.
\newblock \showarticletitle{Comparative Analysis and Design of Cybersecurity
  Maturity Assessment Methodology Using NIST CSF, COBIT, ISO/IEC 27002 and PCI
  DSS}.
\newblock \bibinfo{journal}{\emph{JOIV: International Journal on Informatics
  Visualization}} \bibinfo{volume}{4}, \bibinfo{number}{4}
  (\bibinfo{year}{2020}), \bibinfo{pages}{225--230}.
\newblock


\bibitem[Szegedy et~al\mbox{.}(2013)]%
        {szegedy2013intriguing}
\bibfield{author}{\bibinfo{person}{Christian Szegedy},
  \bibinfo{person}{Wojciech Zaremba}, \bibinfo{person}{Ilya Sutskever},
  \bibinfo{person}{Joan Bruna}, \bibinfo{person}{Dumitru Erhan},
  \bibinfo{person}{Ian Goodfellow}, {and} \bibinfo{person}{Rob Fergus}.}
  \bibinfo{year}{2013}\natexlab{}.
\newblock \showarticletitle{Intriguing properties of neural networks}.
\newblock \bibinfo{journal}{\emph{arXiv preprint arXiv:1312.6199}}
  (\bibinfo{year}{2013}).
\newblock


\bibitem[Vilone and Longo(2021)]%
        {vilone2021notions}
\bibfield{author}{\bibinfo{person}{Giulia Vilone} {and} \bibinfo{person}{Luca
  Longo}.} \bibinfo{year}{2021}\natexlab{}.
\newblock \showarticletitle{Notions of explainability and evaluation approaches
  for explainable artificial intelligence}.
\newblock \bibinfo{journal}{\emph{Information Fusion}} (\bibinfo{year}{2021}).
\newblock


\bibitem[Weidman et~al\mbox{.}(2020)]%
        {weidman2020nothing}
\bibfield{author}{\bibinfo{person}{Jake Weidman}, \bibinfo{person}{Igor
  Bilogrevic}, {and} \bibinfo{person}{Jens Grossklags}.}
  \bibinfo{year}{2020}\natexlab{}.
\newblock \showarticletitle{Nothing Standard About It: An Analysis of Minimum
  Security Standards in Organizations}. In \bibinfo{booktitle}{\emph{European
  Symposium on Research in Computer Security}}. Springer,
  \bibinfo{pages}{263--282}.
\newblock


\bibitem[Wysopal et~al\mbox{.}(2006)]%
        {wysopal2006art}
\bibfield{author}{\bibinfo{person}{Chris Wysopal}, \bibinfo{person}{Lucas
  Nelson}, \bibinfo{person}{Elfriede Dustin}, {and} \bibinfo{person}{Dino
  Dai~Zovi}.} \bibinfo{year}{2006}\natexlab{}.
\newblock \bibinfo{booktitle}{\emph{The art of software security testing:
  identifying software security flaws}}.
\newblock \bibinfo{publisher}{Pearson Education}.
\newblock


\bibitem[Yona and Rothblum(2018)]%
        {yona2018probably}
\bibfield{author}{\bibinfo{person}{Gal Yona} {and} \bibinfo{person}{Guy
  Rothblum}.} \bibinfo{year}{2018}\natexlab{}.
\newblock \showarticletitle{Probably approximately metric-fair learning}. In
  \bibinfo{booktitle}{\emph{International Conference on Machine Learning}}.
  PMLR, \bibinfo{pages}{5680--5688}.
\newblock


\bibitem[Zhang and Shah(2014)]%
        {zhang2014fairness}
\bibfield{author}{\bibinfo{person}{Chongjie Zhang} {and}
  \bibinfo{person}{Julie~A Shah}.} \bibinfo{year}{2014}\natexlab{}.
\newblock \showarticletitle{Fairness in multi-agent sequential
  decision-making}. In \bibinfo{booktitle}{\emph{Advances in Neural Information
  Processing Systems}}. \bibinfo{pages}{2636--2644}.
\newblock


\bibitem[Zhang et~al\mbox{.}(2020)]%
        {zhang2020online}
\bibfield{author}{\bibinfo{person}{Xuezhou Zhang}, \bibinfo{person}{Xiaojin
  Zhu}, {and} \bibinfo{person}{Laurent Lessard}.}
  \bibinfo{year}{2020}\natexlab{}.
\newblock \showarticletitle{Online data poisoning attacks}. In
  \bibinfo{booktitle}{\emph{Learning for Dynamics and Control}}. PMLR,
  \bibinfo{pages}{201--210}.
\newblock


\bibitem[Zhao et~al\mbox{.}(2021)]%
        {zhao2021calibrating}
\bibfield{author}{\bibinfo{person}{Shengjia Zhao}, \bibinfo{person}{Michael
  Kim}, \bibinfo{person}{Roshni Sahoo}, \bibinfo{person}{Tengyu Ma}, {and}
  \bibinfo{person}{Stefano Ermon}.} \bibinfo{year}{2021}\natexlab{}.
\newblock \showarticletitle{Calibrating Predictions to Decisions: A Novel
  Approach to Multi-Class Calibration}.
\newblock \bibinfo{journal}{\emph{Advances in Neural Information Processing
  Systems}}  \bibinfo{volume}{34} (\bibinfo{year}{2021}).
\newblock


\end{thebibliography}

\end{document}